\documentclass[runningheads]{llncs}
\usepackage{graphicx}
\usepackage{todonotes}
\usepackage{mathtools}
\usepackage{multirow}
\usepackage{subfigure}
\usepackage{wrapfig}
\usepackage{url}
\usepackage[inline]{enumitem}
\usepackage{listings}
\usepackage{amsfonts}
\usepackage{amsmath}
\usepackage{mathabx}
\usepackage{lscape}
\usepackage{enumitem}
\usepackage[T1]{fontenc}
\usepackage{verbatim}
\usepackage{wrapfig}
\usepackage{float}
\usepackage{comment}
\usepackage{hyperref}
\usepackage[noend]{algpseudocode}
\usepackage{booktabs}
\usepackage{makecell}
\usepackage{algorithm}
\algrenewcommand{\algorithmiccomment}[1]{$//$ #1}
\newcommand{\orcidJB}	{\href{https://orcid.org/0000-0002-3979-400X}{\protect\includegraphics[scale=0.045]{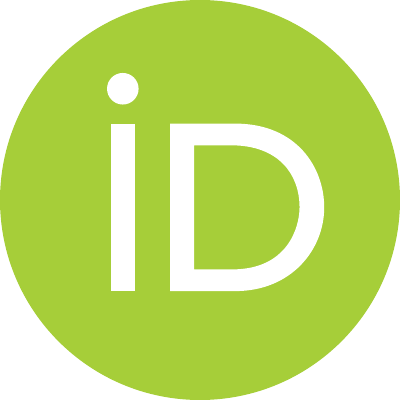}}}
\newcommand{\orcidSRM}	{\href{https://orcid.org/0000-0001-5656-6108}{\protect\includegraphics[scale=0.045]{orcid}}}
\newcommand{\orcidJM}	{\href{https://orcid.org/0000-0002-6332-5801}{\protect\includegraphics[scale=0.045]{orcid}}}
\newcommand{\orcidGP}	{\href{https://orcid.org/0000-0001-9394-6513}{\protect\includegraphics[scale=0.045]{orcid}}}

\makeatletter
\def\namedlabel#1#2{\begingroup
    #2%
    \def\@currentlabel{#2}%
    \phantomsection\label{#1}\endgroup
}
\makeatother

\newcommand{\UEI}{{\mathbb{U}_{ei}}} 
\newcommand{\UACT}{{\mathbb{U}_{act}}} 
\newcommand{\UTIME}{{\mathbb{U}_{time}}} 
\newcommand{\UOT}{{\mathbb{U}_{ot}}} 
\newcommand{\UAOT}{{\mathbb{U}_{aot}}}

\newcommand{\UROT}{{\mathbb{U}_{rot}}} 
\newcommand{\UWFROT}{{\mathbb{U}_{wfot}}}

\newcommand{\UOI}{{\mathbb{U}_{oi}}} 
\newcommand{\UOMAP}{{\mathbb{U}_{omap}}} 
\newcommand{\UE}{{\mathbb{U}_{event}}}

\newcommand{\PI}{\pi_{ei}}
\newcommand{\PA}{\pi_{act}}
\newcommand{\PT}{\pi_{time}}
\newcommand{\PWFO}{\pi_{wfomap}}
\newcommand{\PAO}{\pi_{aomap}}

\newcommand{\MI}{M_{init}}
\newcommand{\MF}{M_{final}}

\newcommand{\OR}{\preceq_E}
\newcommand{\UL}{\mathbb{U}_{OCEL}}

\newcommand{\UAN}{\mathbb{U}_{AN}}

\newcommand{\UP}{\!\!\upharpoonright}

\newcommand{\janik}[1]{\textcolor{black}{#1}}

\newcommand{\final}[1]{\textcolor{black}{#1}}

\begin{document}
\title{INEXA: Interactive and Explainable Process Model Abstraction Through Object-Centric Process Mining
}
\titlerunning{INEXA}
%
\author{Janik-Vasily Benzin\inst{1}\orcidJB \and
Gyunam Park\inst{2}\orcidGP \and
Jürgen Mangler\inst{1}\orcidJM \and 
Stefanie Rinderle-Ma\inst{1}\orcidSRM}
\authorrunning{J.-V. Benzin et al.}
%
\institute{Technical University of Munich, TUM School of Computation, Information and Technology, Garching, Germany \email{\{janik.benzin,juergen.mangler,stefanie.rinderle-ma\}@tum.de} \and
Process and Data Science, RWTH Aachen University, Aachen, Germany
\email{gnpark@pads.rwth-aachen.de}}
\maketitle              
\begin{abstract}
Process events are recorded by multiple information systems at different granularity levels.
Based on the resulting event logs, process models are discovered at different granularity levels, as well. 
Events stored at a fine-grained granularity level, for example, may hinder the discovered process model to be displayed due the high number of resulting model elements. 
The discovered process model of a real-world manufacturing process, for example,
consists of 1,489 model elements and over 2,000 arcs.
Existing process model abstraction techniques could help reducing the size of the model, but would disconnect it from the underlying event log. Existing event abstraction techniques do neither support the analysis of mixed granularity levels, nor interactive exploration of a suitable granularity level. To enable the exploration of discovered process models at different granularity levels, we propose INEXA, an interactive, explainable process model abstraction method that keeps the link to the event log. As a starting point, INEXA aggregates large process models to a ``displayable'' size, e.g., for the manufacturing use case to a process model with 58 model elements. Then, the process analyst can explore granularity levels interactively, while applied abstractions are automatically traced in the event log for explainability. 

\end{abstract}
\section{Introduction}
\label{sec:intro}

Due to the heterogeneity of information systems supporting business processes, e.g., Enterprises Resource Planning and Customer Relationship Management systems, process-related events are often recorded in event logs at different granularity levels. Existing process discovery techniques operate on these event logs and are mostly not able to abstract events for discovery. Hence, \emph{event abstraction} methods have been developed as part of event pre-processing for process mining \cite{van_zelst_event_2021}.The alternative to the event pre-processing for dealing with multiple granularity levels are \emph{process model abstraction} methods that allow to explore granularity levels in the discovered process model by aggregating control-flow structures containing multiple transitions into a single transition.

To illustrate the challenge of multiple granularity levels in process discovery, an event log fragment $L$ is depicted in \autoref{tab:motivating-events-table}. This running example is based on a bank's account opening business process \cite{tsagkani_process_2022}. The two process participants client and bank, are conceptualized as two object types \texttt{workflow:client} and \texttt{workflow:bank}, i.e., the event log is conceptually a \emph{object-centric} event log in which no single case notion is assumed to exist \cite{van_der_aalst_object-centric_2019,ghahfarokhi_ocel_2021}. The recorded events have different granularity levels: the first four events record coarse-grained activities, while the next four events record fine-grained (technical) activities that require aggregation  to achieve a similar granularity level. Moreover, for activity ``finalize account opening''  lifecycle events ``start'', ``on hold'', ``continue'' and ``end'' have been recorded, resulting in the last four events at different granularity level.

Event abstraction works by aggregating fine-granular events into a single coarse-grained event, e.g., the last four events in \autoref{tab:motivating-events-table} would be aggregated into a single event with the activity label ``finalize account opening''. Then, process discovery technique $pd$ can discover a process model \cite{beerepoot_biggest_2023}. Despite advancements in event abstraction methods three challenges remain. 

\begin{figure}
  \centering
  \includegraphics[width=\linewidth]{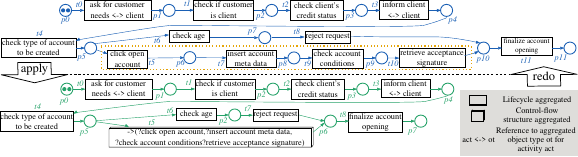}
  \caption{Two process models of a bank's account opening business process \cite{tsagkani_process_2022} abstracted with INEXA. The upper process model shows four transitions with fine-grained activity labels denoted in orange, while the lower process model exhibits only a single transition $t5$ as an aggregate of the four transitions with fine-grained activity labels.}
  \label{fig:running}
\end{figure}

First, once an abstraction level is determined by an event abstraction method, all the following discovery and analysis steps are bound to the abstraction level, e.g., transition $t11$ cannot be explored in detail in the analysis phase in the upper process model in \autoref{fig:running}. 
Second, the event abstraction does not offer any functionality of directly abstracting patterns in the process model that are too fine-grained, i.e., the analyst has to first identify the corresponding events in the log and a suitable event abstraction, followed by applying process discovery another time. Third, the event abstraction does not trace the analysis journey, i.e., the iterations through abstraction, discovery and analysis, such that process analysts can have difficulties to explain how their analysis resulted in the final process model \cite{zerbato_granularity_2021,beerepoot_biggest_2023}. Event abstraction supports neither \emph{interactive} exploration without iterating through multiple methods due to the first two challenges nor \emph{explainability} of the abstracted process model due to the third challenge.

\begin{table}[htb!]
    \centering
    \caption{Event log fragment $L$ based on \cite{tsagkani_process_2022}. Each event can refer to objects of a certain object type (columns workflow:client to abstraction:workflow:client\$caa are object types). An event is represented by a row (except the header). The two object type columns in blue are the result of two augmented event log transitions $st_{abs_{cla}}, st_{abs_{caa}}$ that augment the original event log with abstraction objects.}
    \scalebox{0.65}{
    \begin{tabular}{lccccccc} 
    \toprule
    id & activity    & timestamp           & \makecell{workflow:\\client}                                     & \makecell{workflow:\\bank} &  \makecell{workflow:lc:\\finalize\\account\\opening}   & \makecell{\textcolor{blue}{abstraction:}\\\textcolor{blue}{workflow:lc:}\\\textcolor{blue}{finalize}\\\textcolor{blue}{account}\\\textcolor{blue}{opening\$cla}} & \makecell{\textcolor{blue}{abstraction:}\\\textcolor{blue}{workflow:}\\\textcolor{blue}{client\$caa}}\\ 
    \midrule
    0ab63  & ask for customer needs & 2023-05-19T10:42:49 & \{a0287\}      &  \{151a3\}    &   $\emptyset$   &   \textcolor{blue}{$\emptyset$} &   \textcolor{blue}{\{uih13\}} \\
    6b0b9 & check if customer is client & 2023-05-19T10:43:36 &       $\emptyset$               &  \{151a3\}        & $\emptyset$  &   \textcolor{blue}{$\emptyset$} &   \textcolor{blue}{$\emptyset$}     \\
     ddf21 & check client's credit status & 2023-05-19T10:44:25 &    $\emptyset$                &   \{151a3\}   &  $\emptyset$  &   \textcolor{blue}{$\emptyset$} &   \textcolor{blue}{$\emptyset$}    \\
    9c7f8 & inform client & 2023-05-19T10:45:16 &      \{a0287\}               &  \{151a3\}         & $\emptyset$  &   \textcolor{blue}{$\emptyset$} &   \textcolor{blue}{\{uih13\}}   \\
    207g2 & \makecell{check type of account\\to be created} & 2023-05-19T10:55:29 &      \{a0287\}               &  \{151a3\}         & $\emptyset$  &   \textcolor{blue}{$\emptyset$} &   \textcolor{blue}{\{uih13\}}  \\
    260f5 & click open account & 2023-05-19T10:57:25 &      $\emptyset$                &  \{151a3\}         & $\emptyset$   &   \textcolor{blue}{$\emptyset$} &   \textcolor{blue}{$\emptyset$}     \\
    0a1e4 & insert account meta data & 2023-05-19T10:57:27 &      $\emptyset$              &  \{151a3\}         & $\emptyset$   &   \textcolor{blue}{$\emptyset$} &   \textcolor{blue}{$\emptyset$}    \\
    6629a & check account conditions & 2023-05-19T10:58:03 &      $\emptyset$               &  \{151a3\}         & $\emptyset$    &   \textcolor{blue}{$\emptyset$} &   \textcolor{blue}{$\emptyset$}    \\
    1c0bf & retrieve acceptance signature & 2023-05-19T11:58:59 &      $\emptyset$                &  \{151a3\}         & $\emptyset$   &   \textcolor{blue}{$\emptyset$} &   \textcolor{blue}{$\emptyset$}      \\
    48c83 & \makecell{finalize account opening\\- start} & 2023-05-19T11:00:33 &      $\emptyset$            &     \{151a3\}       & \{b8955\}   &   \textcolor{blue}{\{kl273\}} &   \textcolor{blue}{$\emptyset$}     \\
    ddf21 & \makecell{finalize account opening\\- on hold} & 2023-05-19T11:00:45 &      $\emptyset$            &     \{151a3\}       & \{b8955\}   &   \textcolor{blue}{\{kl273\}} &   \textcolor{blue}{$\emptyset$}     \\
    kj875 & \makecell{finalize account opening\\- continue} & 2023-05-19T11:01:51 &      $\emptyset$            &     \{151a3\}      & \{b8955\}   &   \textcolor{blue}{\{kl273\}} &   \textcolor{blue}{$\emptyset$}     \\
    87bd9 & \makecell{finalize account opening\\- end} & 2023-05-19T11:02:12 &      $\emptyset$            &    \{151a3\}       & \{b8955\}   &   \textcolor{blue}{\{kl273\}} &   \textcolor{blue}{$\emptyset$}     \\
    \bottomrule
    \end{tabular}}
    \label{tab:motivating-events-table}
\end{table}

Process model abstraction abstracts control-flow patterns in the process model instead of events \cite{smirnov_business_2012,senderovich_aggregate_2018,tsagkani_process_2022}. 
Existing methods do not fully support interactive exploration of different granularity levels \final{($\rightarrow$ limited interactivity)} and only trace the analysis journey in a restricted fashion. This is because neither can the process analyst explain what events were abstracted to yield the abstracted process model nor is the sequence of applied abstractions during exploration stored in a standardized format with the process model \final{($\rightarrow$ limited explainability)}. Moreover, event logs cannot be \emph{replayed} on abstracted process model, i.e., subsequent analysis is limited. 

Overall, the lack of existing methods for interactive, explainable process model abstraction raises the following research question:

\noindent\textbf{RQ} How can we construct a process model abstraction method that allows to interactively explore granularity levels in the discovered process model, while tracing the model abstraction in the corresponding event log for explainability?

To address \textbf{RQ} we propose the INEXA method for interactive, explainable process model abstraction that overcomes existing challenges for event abstraction and process model abstraction. INEXA first discovers a process model on the original event log. Then, it applies model abstractions to the process model. For example, the \emph{complete lifecycle} abstraction (cla) for activity ``finalize account opening'' that aggregates the four transitions labelled with the activity and its lifecycle is applied by INEXA such that transition $t11$ in the upper process model in \autoref{fig:running} is the aggregate of the activity's lifecycle model. Also, the \emph{complete artifact} abstraction (caa) that completely aggregates all model elements pertaining to a process \emph{artifact}, i.e., object types that conceptualize process artifacts with a workflow as object types, is applied by INEXA for the client artifact such that only a label reference ``$\leftrightarrow$ client'' in affected transitions of the bank are kept \final{to focus on the bank similar to \cite{tsagkani_process_2022}} (cf. transition $t0$ and $t3$ in the upper process model in \autoref{fig:running}).

Instead of abstracting the events corresponding to the activity lifecycle or the client, INEXA adds an abstraction object ``kl273'' corresponding to the ``cla'' abstraction and abstraction object ``uih13'' corresponding to the client ``caa'' abstraction (cf. blue columns in \autoref{tab:motivating-events-table}). The object type of abstraction objects and related events are sufficient to project the originally discovered process model to the abstracted process model through a novel \emph{log-model link}. From an (abstracted) process model, INEXA allows the process analyst to apply further abstractions, e.g., abstracting the sequence of fine-grained activities corresponding to the four events with ids ``260f5'', ``0a1e4'', ``6629a'' and ``1c0bf'' resulting in the further abstracted process model on the bottom. For interaction, INEXA allows to redo already applied abstractions, e.g., the process analyst can redo the most recent applied sequence abstraction (cf. \autoref{fig:running}). Both operations are traced by abstraction objects and an abstraction history object in the event log.

To enable the log-model link, INEXA builds on \emph{object-centric} process mining (OCPM), a new paradigm of process mining methods that does not assume a single case notion in the event log. Event logs in OCPM consist of events that are related to objects of certain object types, a property that is exploited by INEXA by adding further abstraction objects that are interpreted as abstraction views on the process. By distinguishing classes of object types, e.g., \emph{workflow} types from \emph{abstraction} types (cf. first three object type columns vs. last two columns in \autoref{tab:motivating-events-table}), the original event log remains unchanged, while INEXA augments the event log by adding objects of abstraction types.

The remainder of this paper is structured as follows. \autoref{sec:disc} introduces a taxonomy of object types that is required by INEXA. \autoref{sec:method} proposes INEXA. i.e., idea, concepts, an initial set of process model abstractions, and the overall method. INEXA is evaluated based on a real-world dataset from the manufacturing domain in \autoref{sec:eval}. Next, \autoref{sec:rel} summarizes related work. Lastly, \autoref{sec:con} concludes the paper.

\section{Object Type Taxonomy in Object-centric Event Logs}
\label{sec:disc}
This section proposes a taxonomy of object types that differentiates two fundamentally different families of object type classes: \emph{record} object types and \emph{abstraction} object types in \autoref{ssec:ot}. Given the taxonomy of object types, we state concepts and notions of object-centric process mining and process model abstraction in \autoref{ssec:prelim}.

\subsection{Workflow Object Types vs. Abstraction Object Types}
\label{ssec:ot}

\begin{figure}[htb!]
  \centering
  \includegraphics[width=\linewidth]{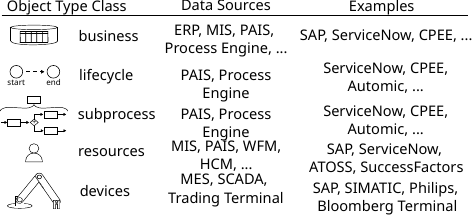}
  \caption{Workflow object type classes conceptualized with the aim of representing these artifacts based on   \cite{russell2004workflow,buhner_working_2006,reijers_ha_usefulness_2010,smirnov_business_2012,dumas_fundamentals_2013,mangler_cpee_2014,kumar_optimal_2013,pika_mining_2017,van_der_aalst_object-centric_2019,van_der_aalst_discovering_2020,turetken_influence_2020,ghahfarokhi_ocel_2021,ehrendorfer_assessing_2021,park2022detecting,adams_defining_2022,fdhila_verifying_2022,tsagkani_process_2022}.}
  \label{fig:rtypes}
\end{figure}

Object types in \emph{object-centric event logs} (cf. \autoref{tab:motivating-events-table}) and process mining are commonly conceptualized as types of business objects, e.g., orders, items and packages recorded in an information system \cite{van_der_aalst_object-centric_2019,van_der_aalst_discovering_2020,ghahfarokhi_ocel_2021,adams_defining_2022}. Hence, objects have a workflow that can be discovered by \emph{object-centric process discovery} (OCPD) directly on the event log for theses common object types that can be extracted from ERP, e.g., SAP, and related data sources (cf. \autoref{fig:rtypes}).

Business object types have a workflow similar to the \emph{lifecycle}, \emph{subprocess}, \emph{resources} and \emph{devices} object types classes as depicted in \autoref{fig:rtypes}. Business object types can be accompanied by the lifecycle and subprocess object types to support representing these concepts from process-aware information systems (PAIS) \cite{dumas_fundamentals_2013} or process engines \cite{mangler_cpee_2014}. Conceptualizing the activity lifecycle as an object type allows to discover activity lifecycle models without further changes to the discovery method as, e.g., proposed by \cite{reichert_using_2016} as a lifecycle extension to the IM discovery technique.

Resources and devices in the business process have their own workflow corresponding to the daily work schedule of a particular resource \cite{russell2004workflow,bertrand_survey_2023} or device. Both are typically extracted as event attributes in process mining \cite{van_der_aalst_process_2016}. However, a object type extraction supports integrating resource and device event information from specialized data sources such as human capital management (HCM) systems or device logs into a single event log to support further analysis, e.g., resource- and device-based views for abstraction \cite{smirnov_business_2012}.

\begin{figure}
  \centering
  \includegraphics[width=\linewidth]{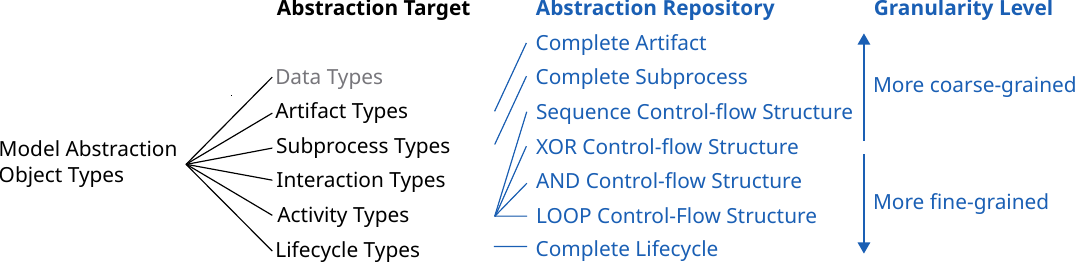}
  \caption{Model abstraction object type classes classified by ``abstraction target'' based on \cite{smirnov_business_2012,tsagkani_process_2022}. All types in black in column ``abstraction target'' can be applied in INEXA. INEXA's initial \emph{abstraction repository} (cf. \autoref{ssec:prelim}) consists of the seven process model abstraction instances in column ``abstraction repository'' from coarse-grained to fine-grained granularity level.}
  \label{fig:atypes}
\end{figure}

The first three columns of object types in \autoref{tab:motivating-events-table} are workflow object types. 
Workflow object types are extracted from the various data sources of a company. A distinction into the eight classes enables INEXA to discover a process model for artifacts having a workflow, supports the selection of resource- and machine-based viewpoints with a similar mechanism as for business objects, facilitates further analysis with respect to interdependencies uncovered by integrating multiple data sources during extraction and constitutes a direct interface to lifecycle and subprocess abstractions.
In contrast to the workflow object types, abstraction object types (cf. \autoref{fig:atypes}) are introduced by INEXA to trace the process model abstraction in the event log and have to be ignored by OCPD.

The abstraction object types are not recorded in information systems and additional data sources (cf. \autoref{fig:rtypes}), but are added to the event log during process model abstraction in INEXA (cf. \autoref{ssec:method}). Therefore, workflow object types and objects are not changed by model abstraction, i.e., the original, extracted event log is maintained without changes with respect to workflow object types. \final{\cite{smirnov_business_2012} review the process model abstraction literature and identify \emph{aggregation} as the abstraction that affects the granularity level.} As aggregation takes multiple abstraction targets and summarizes them into a single, more coarse-granular element and, thus, moves the model to a more coarse-grained granularity level \cite{smirnov_business_2012}, aggregation is the type of abstraction relevant to INEXA (cf. \autoref{sec:intro}). Hence, all abstractions in \autoref{fig:atypes} are aggregations.

Based on the use cases in \cite{smirnov_business_2012,tsagkani_process_2022}, we distinguish six abstraction targets for aggregation. \final{The artifact type subsumes all abstractions that affect the whole process artifact, e.g., the complete artifact abstraction for the client aggregates the client completely in the running example (cf. \autoref{tab:motivating-events-table}). The subprocess type subsumes abstractions that are aggregating subprocess workflows that are at a more fine-grained granularity levels, since subprocesses modularize parent processes. For example, subprocess workflow objects can be directly used to aggregate the complete subprocess (cf. \autoref{fig:atypes}) as demonstrated in \autoref{sec:eval}. Interaction types aggregate \emph{artifact interactions}, e.g., the two client interactions could be aggregated to a single client interaction with the bank in \autoref{fig:running}. Interaction types are not part of the \emph{abstraction repository} due to the assumption that interactions are significant \cite{tsagkani_process_2022}. Activity types are all aggregations that aggregate a subset of transitions of a certain process artifact that represent a certain control-flow pattern, e.g., the sequence of transitions $t5$, $t7$, $t9$ and $t10$ in the upper process model in \autoref{fig:running}. The most fine-grained abstraction target is the lifecycle of a single activity, e.g., aggregating all transitions with a lifecycle event of activity ``finalize account opening'' in \autoref{tab:motivating-events-table} with a ``cla'' abstraction.} 

\final{An abstraction is \emph{transition-reconstructable} if it can be uniquely reconstruct\-ed by its type that references the abstraction's definition in the abstraction repository and by a set of transitions that it aggregates, e.g., the activity type ``Sequence control-flow structure'' and transitions $t5$, $t7$, $t9$ and $t10$ are sufficient to abstract the upper process model to the lower process model in \autoref{fig:running}. Abstractions that are not transition-reconstructable are not supported by INEXA (marked in grey in Fig. 3).}
To formalize workflow and supported abstraction object type classes, we distinguish these classes of object types as follows.

\begin{definition}[Workflow, Model Abstraction and Abstraction History Object Type(s)]
\label{def:atypes}
 Let $\UOT$ be an object type universe, $\UOI$ be a object identifier universe and $type \in \UOI \rightarrow \UOT$ a function mapping object identifiers to their object type. \final{$\UWFROT$ is the workflow object type universe with $\UWFROT \subseteq \UOT$} and all classes in \autoref{fig:rtypes} are in $\UWFROT$. \final{$\UAOT $ is the transition-reconstructable model abstraction object type universe with $\UAOT \subseteq \UOT $ and $\UAOT \cap \UWFROT = \emptyset$}, and all supported classes in \autoref{fig:atypes} are in $\UAOT$. 
 To trace the abstraction history during interactive abstraction, \final{we define the object $absHistory$ of type $history \in \UOT \,\setminus\, \UWFROT \,\setminus\, \UAOT$ with object value $absHistory.applied \in \{aoi \in \UOI \,|\, type(aoi) \in \UAOT \}^*$.} 
\end{definition}

Overall, abstraction object types correspond to process model abstractions, are added to the event log by INEXA through event log transitions (cf. \autoref{alg:st}) and must not be interpreted as workflow object types with respect to OCPD.

\subsection{Preliminaries}
\label{ssec:prelim}

Business process executions are recorded in multiple information systems and extracted into an event log.

\begin{definition}[Event Log]
\label{def:log}
\final{Let $\UEI$, $\UACT$ , $\UTIME$, $\UOMAP$ be the universe of event identifiers, activity labels, timestamps, and object mappings $\{ omap \in \mathbb{U}_{ot} \not\rightarrow \mathcal{P}(\mathbb{U}_{oi}) \,|\, \forall_{ ot \in dom(omap)} \forall_{ oi \in omap(ot)} type(oi) = ot \}$. An event log $L = (E, \OR , \\absHistory)$\footnote{Note that the event log can be stored in the OCEL format \cite{ghahfarokhi_ocel_2021}.} is a subset of events $ E \subseteq \UE = \UEI \times \UACT \times \UTIME \times \mathbb{U}_{wfomap} \times \mathbb{U}_{aomap}$ with $\mathbb{U}_{wfomap} = \{wfomap \in \UOMAP \,|\, dom(wfomap) \subseteq \UWFROT\}$ and $\mathbb{U}_{aomap}$ similarly, a partial order $\OR$ on events and the abstraction history object $absHistory$.}
 We denote the set of event logs as $\mathbb{U}_{OCEL}$.
\end{definition}

\final{For example, the first row of \autoref{tab:motivating-events-table} describes event $e_0$ with $ \PI(e_0) = \text{0ab63}, \PA(e_0) = \text{``ask for customer needs''}, \PT(e_0) = \text{2023-05-19T10:42:49}, \\\PWFO(e_0)(\text{workflow:\\client}) = \{\text{151a3}\}, \ldots$, and $\\\PAO(e_0)(\text{abstraction:workflow:client\$caa}) = \{\text{uih13}\}$\footnote{We use $\pi$ to denote selection of an element/set/function from a tuple.}.}

Business process executions are modelled as a \emph{labeled Petri net}. We use common notation, definitions, and semantics for labeled Petri nets, \emph{accepting Petri nets}, \emph{workflow nets} and \emph{soundness} of workflow nets in the following \cite{van_der_aalst_discovering_2020,van_der_aalst_soundness_2011,benzin_preventing_2023}. Note that we omit (parts of) formal definitions, which are not necessary for subsequent sections, but refer to the respective article for details. 

A process model for an event log is a accepting \emph{object-centric} Petri net. An \emph{object-centric Petri net} is a tuple $ON = (N, pt, F_{var})$ where $N = (P, T, F, L)$ is a labeled Petri net, $pt \in P \rightarrow \UWFROT$ maps places onto object types, and $F_{var} \subseteq F$ is the subset of variable arcs. In object-centric Petri nets, the respective object execution flows of a certain type are differentiated by typing the Petri net's places with object types (cf. bank object type denoted in blue in \autoref{fig:running}). We identify the process model for a workflow object type $ot$ in $ON$ as $ON\UP_{ot} = APN^{ot} = (N^{ot}, T^{ot}, F^{ot}, l\UP_{T^{ot}})$ with $N^{ot} = \{p \in P | pt(p) = ot \}$, $T^{ot} = \{ t \in T | \exists_{p \in \bullet t \cup t \bullet} pt(p) = ot\}$, and $F^{ot} = F \cap ((P^{ot} \times T^{ot}) \cup (T^{ot} \times P^{ot}))$\footnote{Given $X$, we denote the restriction of function $f$ on $X'$ as $f\UP_{X'} = \{ (x', f(x')) | x' \in X' \}$.}. Adding an initial $\MI$ and final marking $\MF$ to an object-centric Petric net defines an accepting object-centric Petri net $AN = (ON, \MI, \MF) $ \cite{van_der_aalst_discovering_2020}. We denote the universe of accepting object-centric Petri nets as $\UAN$. The event log $L = (E, \OR , ovmap)$ \emph{perfectly fits} accepting object-centric Petri net $AN$ iff all events $e \in E$ can be \emph{replayed} on $AN$ \cite{aalst_replaying_2012} and correspond to a firing sequence from the initial to the final marking.
We also write $\MI \xrightarrow{L} \MF$, if event log $L$ perfectly fits process model $AN$.

A object-centric process discovery (OCPD) technique $pd \in \UL \rightarrow \mathbb{U}_{AN}$ discovers a process model for an event log. If the Inductive Miner (IM) discovery technique \cite{leemans_discovering_2013} is used within OCPD, we denote the resulting technique as $pd_{IM}$ \cite{van_der_aalst_discovering_2020}. Since events $e$ can be related to multiple objects $oi_1, oi_2, \ldots \in \UOI$ of multiple types $ot_1, ot_2, \ldots \in \UWFROT$ (cf. \autoref{tab:motivating-events-table}), a OCPD technique discovers \emph{interaction transitions} $t \in T$ with correspondingly typed places $pt(p_1) = pt(p_2) = ot_1, pt(p_3) = pt(p_4) = ot_2, \ldots$ in the pre-set $p_1, p_3, \ldots \in \bullet t$ and post-set $p_2, p_4, \ldots \in t \bullet$, i.e., interaction transitions model the synchronous interaction of multiple object executions \cite{van_der_aalst_process_2016,van_der_aalst_discovering_2020,benzin_preventing_2023}.

Process models can be abstracted with existing model abstractions \cite{smirnov_business_2012,senderovich_aggregate_2018,tsagkani_process_2022}.

\begin{definition}[Model Abstraction Repository and Admissibility]
\label{def:abs}
 A mo\-del abstraction is a function $abs: \UAN \rightarrow \UAN $, such that for all $AN \in \UAN$ it holds that $|abs(AN)| \leq |AN|$ \cite{senderovich_aggregate_2018}. A model abstraction $abs$ has a type $type(abs) \in \UAOT$\footnote{We overload the function $type$.}. An abstraction repository is a function $ar \in \UAOT \rightarrow (\UAN \rightarrow \UAN)$ that maps abstraction types $type(abs)$ to their corresponding model abstractions $abs$.  
\end{definition}

In the running example (cf. \autoref{fig:running}), a model abstraction $abs_{\text{sequence}}$ aggregates the sequence of transitions $T' = \{ t5, t7, t9, t10\}$ labelled with fine-grained activities ``click open account'' to ``retrieve acceptance signature'' into a single transition $t_5$ labelled with activity ``$\rightarrow$(?click open account, ..., ?retrieve acceptance signature)'', i.e., the transition set $T'$ and its type ``Sequence control-flow structure'' (cf. \autoref{fig:atypes}) are sufficient to reconstruct the concrete model abstraction $abs_{\text{sequence}}$ that is applied in \autoref{fig:running}. 

\final{We subsume the conditions that need to be satisfied for a specific model abstraction $abs$ to be applicable for a set of transitions $T$ in a predicate $adm_{T} \in (\UAN \rightarrow \UAN) \rightarrow \mathbb{B}$. For the abstraction $abs_{\text{sequence}}$ with the set of transitions T' in the upper process model in \autoref{fig:running}, the predicate $adm_{T'}(abs_{\text{sequence}})$ evaluates to True. First, the abstraction is \emph{order-preserving} \cite{liu_workflow_2003,polyvyanyy2009application,bobrik_view-based_2007,smirnov_business_2012}, since it does not reverse any order constraints imposed by places $P$ in the lower process model in \autoref{fig:running} compared to the upper process model. Second, it is \emph{soundness-preserving} \cite{polyvyanyy2009application,smirnov2012business}, a property that guarantees soundness of the lower process model, if the upper process model is sound in \autoref{fig:running}. Third, the set of transitions T' is a single-entry/single-exit SESE control-flow structure \cite{bobrik_view-based_2007,vanhatalo_refined_2009,polyvyanyy_simplified_2011} of type ``Sequence''. Note that given the technique proposed in \cite{van_zelst_translating_2020}, we can determine with $isSESE_{aot} \in T \rightarrow \mathbb{B}$ whether a subset of transitions $T'' \subseteq T$ of a process model is a SESE control-flow structure of type $aot \in \{\text{Sequence}, \text{XOR}, \text{AND}, \text{LOOP}\}$, e.g., $isSESE_{\text{sequence}}(T') = \text{True}$. These SESE control-flow structures are aggregated by their respective abstractions. We refer to \cite{senderovich_aggregate_2018,van_zelst_translating_2020} for formal definitions of the model abstractions ``Sequence'', ``XOR'', ``AND'', and ``LOOP control-flow structure'' in the activity aggregation class.}

Model abstraction types (cf. \autoref{ssec:ot}) can be encoded into an abstraction object with the corresponding object type (cf. types in black in \autoref{fig:atypes}) that are added to the augmented event log and traced in the abstraction history object by the corresponding event log transition $st_{abs}$ (cf. \autoref{alg:st}). Thus, an event log is \emph{augmented}, if the abstraction history $absHistory$'s value $applied$ is not empty. \final{$L\UP_{\UWFROT}$ denotes the event log without the abstraction object map $aomap$ in each event.}

Since workflow object types model various workflows of business objects, resources, devices, subprocesses and lifecycle (cf. \autoref{ssec:ot}), INEXA also supports artifact (for business objects, resources and devices), subprocess and lifecycle aggregation in the form of completely aggregating all model elements for that workflow object type. Hence, the three model abstractions ``complete artifact'', ``complete subprocess'', and ``complete lifecycle'' are completely aggregating the corresponding workflow object type $ot \in \UWFROT$ as if $ot$ was not part of the event log, but keeps references in the activity label of interacting transitions as depicted in \autoref{fig:running}. Following the assumption of \cite{tsagkani_process_2022} that \emph{interactions} are significant and should not be abstracted, we do not demonstrate the interaction aggregation in INEXA. As these interactions are modelled by interaction transitions, they are, in principle, supported by INEXA. 

\section{INEXA: From Concepts to the Method}
\label{sec:method}

This section presents the INEXA method by 
presenting the interactive interface based on the log-model link in \autoref{ssec:idea} and specifying the method in \autoref{ssec:method}. 

\subsection{INEXA's Interactive Interface based on the Log-Model Link}
\label{ssec:idea}

\begin{figure}
  \centering
  \includegraphics[width=0.7\linewidth]{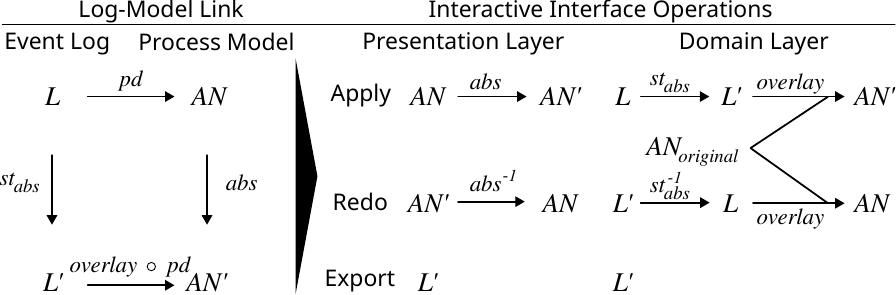}
  \caption{General idea for interactive, explainable model abstraction INEXA with separated presentation and domain layer for operations based on Domain-Driven Design \cite{evans2004domain}. $pd$ is an object-centric process discovery technique, $abs$ a process model abstraction, $st_{abs}$ the corresponding augmented event log transition, and $overlay$ the projection of an augmented event log $L$ onto the original process model $AN_{original}$.}
  \label{fig:link}
\end{figure}

While model abstraction is presented to the user for interactive and usually more understandable abstraction functionality \cite{zerbato_granularity_2021,tsagkani_process_2022}, the corresponding event log transition is developed to overcome the challenges of maintaining the ability to replay the event log on the process model and tracing the analysis journey. As depicted in \autoref{fig:link}, the conceptual log-model link targets the challenge of linking model and event log after model abstractions are applied to the process model. It is established by mirroring model abstraction $abs$ on process models through event log transitions $st_{abs}$ on the corresponding augmented event logs. 
Then, \emph{model overlay} function $overlay$ completes the log-model link by taking the augmented event log $L'$ and the non-abstracted process model $AN$ discovered by $pd$ and maps it to the abstracted process model $AN'$. Hence, the log-model link is defined by an event log state transition $st_{abs}$, followed by the discovery of the original process model $AN$ through $pd$ and overlaying the original model with the augmented event log $L'$ to yield the abstracted process model $AN'$.

The interactive interface features \janik{a minimal set of functional operations required for interaction}: \texttt{Apply}, \texttt{redo} and \texttt{export}. All three operations are presented in \autoref{ssec:method} in detail, since they build on the log-model link's functions $st_{abs}$ and $overlay$. \final{The \texttt{apply} operation enables the application of model abstraction $abs$ to the (discovered) process model $AN$ in the presentation layer, while the domain layer does not implement the process model abstraction $abs$ directly on process models, but instead the corresponding augmented event log transition $st_{abs}$ on event logs. The \texttt{redo} operation enables retrieving a previous process model $AN$ from the current process model $AN'$ in which a selected abstraction is reversed in the presentation layer, i.e., the inverse $abs^{-1}$ is applied. In the domain layer, the operation is not achieved by storing the process model $AN$ before the to be redone abstraction was applied, but instead the augmented event log transition $st_{abs}$ is invertible such that the last augmented event log $L$ can be computed. Both operations together enable interactively exploring process model granularity levels by applying multiple abstractions after each other. Operation \texttt{export} presents the current augmented event log $L'$ that contains the history of applied abstractions in $absHistory$. To support quick exploration of granularity levels, we additionally propose operation \texttt{initialize} that is executed once before INEXA offers the minimal set of operations for interaction. \texttt{Initialize} applies a sequence of abstractions until a goal is achieved, e.g., until the abstracted process model reaches a given size.}

\setlength{\textfloatsep}{0pt}
\begin{algorithm}[t]
\scriptsize
\caption{Event Log Transition $st_{abs}$}
\begin{algorithmic}[1]
\Require Event log $L \in \UL$, workflow object type $ot \in \UWFROT$, model abstraction $abs \in \UAN \rightarrow \UAN$ applied on transitions $T'$ of process model $AN\UP_{ot}$ 
\Ensure Augmented event log $L \in \UL$
\State Compute transition to event relation $TE \subseteq T \times E$ through event to transition function $et \in E \rightarrow T$ computed by replaying $L\UP_{\UROT}$ on $AN$
\State \Comment{$EV$ contains events that are aggregated by the abstraction, i.e., these events must be related}
\State \Comment{to the new abstraction object $aoi$ of encoded abstraction object type $eaot$.}
\State $EV \gets \{ e \in E \,| \,\exists_{t \in T'} (t, e) \in TE \}$
\State $aoi \gets generateId()$
\State $eaot \gets encodeAbstractionType(aot)$
\For{$e \in EV$} 
\State \Comment{$f \oplus (x, y) = f'$ is a function operation such that $f'(x) = y, f'(z) = f(z) $ for}
\State \Comment{$z \in dom(f) \setminus \{x\}$.}
\State $\PAO(e) \gets \PAO(e) \oplus (eaot, aoi) $   
\EndFor
\State $absHistory.applied \gets concatenate(absHistory.applied, aoi)$ 
\State \Return $L = (E, \OR, absHistory)$
\end{algorithmic}
\label{alg:st}
\end{algorithm}

In \autoref{alg:st}, the event log transition $st_{abs}$ is presented. The event log transition $st_{abs}$ augments the event log with an abstraction object identifier $aoi$, e.g. ``kl273'' in \autoref{tab:motivating-events-table} for $st_{abs_{cla}}$, and adds that abstraction object $aoi$ to each event $e \in EV$ that is related to an abstracted transition $t \in T'$, e.g., event ``48c83'' in \autoref{tab:motivating-events-table}.

Given an augmented event log $L \in \UL$ and a repository of model abstractions $ar \in \UAOT \rightarrow \UAN \rightarrow \UAN$, the abstracted process model $AN'$ can be reconstructed by model overlay (cf. \autoref{alg:overlay}). 
Model overlay requires a process model $AN$ such that the event log $L\UP_{\UROT}$ perfectly fits $AN$. A perfect fitness is guaranteed by multiple process discovery techniques, e.g., IM  \cite{leemans_discovering_2013} or language-based region miners \cite{bergenthum_process_2007,bergenthum_synthesis_2009}, that can be used for OCPD $pd$, but do not guarantee perfect fitness in object-centric settings \cite{benzin_preventing_2023}. However, perfect fitness can be easily checked together with the requirement that each transition $t \in T$ of the process model occurs at least once in the firing sequence as part of INEXA in \autoref{ssec:method}. Model overlay extracts the sequence of applied model abstractions from the model abstraction history object $history$ by querying the corresponding abstraction type $type(aoi)$ of the current abstraction object $aoi$ from the abstraction repository $ar$, determining the set of transitions $T'$ through the events $E'$ related to the current abstraction object $aoi$ and applying the model abstraction $type(aoi)_{T'}$ on process model $AN$. 

\setlength{\textfloatsep}{0pt}
\begin{algorithm}[t]
\scriptsize
\caption{Model overlay $overlay$}
\begin{algorithmic}[1]
\Require Augmented event log $L \in \UL$, Process model $pd(L\UP_{\UWFROT}) = AN \in \UAN$ that perfectly fits $L\UP_{\UWFROT}$ and the firing sequence contains each transition $t \in T$ at least once, abstraction repository $ar \in \UAOT \rightarrow \UAN \rightarrow \UAN$
\Ensure $AN' = abs^n(AN)$ for $abs^i \in AR, i \in \{0, \ldots, n\}$
\State Compute event to transition function $et \in E \rightarrow T$ by replaying $L\UP_{\UROT}$ on $AN$
\For{$aoi \in absHistory.applied $} 
\State \Comment{Reconstruct abstraction through events $E'$ related to abstraction object $aoi$, transitions}
\State \Comment{$T'$ of $AN$ related to events $E'$, querying abstraction repository $ar$ for abstraction definition}
\State \Comment{and applying retrieved abstraction on the transitions $T'$.}
\State $E' \gets \{ e \in E \;|\; aoi \in \PAO(e)(aot)\}$ \State $T' \gets et(E')$ 
\State $AN \gets ar(type(aoi))_{T'}(AN)$ 
\EndFor
\State \Return $AN$
\end{algorithmic}
\label{alg:overlay}
\end{algorithm}

Given the log-model link, the INEXA method in the next section supports the three operations \texttt{apply}, \texttt{redo} and \texttt{export} as introduced in the last section. 

\subsection{INEXA Method}
\label{ssec:method}

This section specifies the INEXA method for interactive, explainable model abstraction. 
The INEXA method comes with seven aggregations that are denoted in blue in \autoref{fig:atypes} and that define the initial abstraction repository $ar$, i.e., $ar$ is defined for these seven model abstraction types (cf. \autoref{ssec:prelim}).

While the \texttt{export} operation is trivial, as it is a straightforward exposition of the current augmented event log, the \texttt{apply} and \texttt{redo} require a starting point from which interactive exploration of the granularity level can occur. \janik{To that end, the \texttt{initialize} operation of INEXA (cf. \autoref{alg:init}) discovers the original process model $AN$, computes an abstraction tree $AT$, and applies abstractions from the tree as long as the model size exceeds the parameter $sizeThreshold$.}

To minimize the information loss through abstraction, the operation \texttt{init\-ialize} applies aggregations\footnote{\janik{Note that INEXA aims at granularity level such that the abstraction repository only contains abstractions that are aggregations.}} in the order from the most fine-grained to the most coarse-grained granularity level. To that end, the operation computes an abstraction tree \janik{(line 3 to line 16 in \autoref{alg:init}) that has at its root the most coarse-grained granularity level aggregation in the class artifact abstraction (cf. granularity levels of the seven model abstractions $abs$ contained in the abstraction repository $ar$ of INEXA in \autoref{fig:atypes}) and contains further control-flow structure aggregations in the order of the respective workflow objects's process tree}. 

\setlength{\textfloatsep}{10pt}
\begin{algorithm}[t]
\scriptsize
\caption{Initialize}
\label{alg:init}
\begin{algorithmic}[1]
\Require Event log $L \in \UL$ with object types $WFOT \subseteq \UWFROT$, abstraction repository $ar \in \UAOT \rightarrow \UAN \rightarrow \UAN$
\Ensure Abstracted process model $AN' \in \UAN$ with augmented event log $L \in \UL$, abstraction tree $AT$ 
\State $AN = (ON, \MI, \MF) \gets pd_{IM}(L\UP_{\UWFROT})$ 
\If{$\MI \xrightarrow{L} \MF$} 
\For{$ot \in WFOT$}
\State \Comment{Root of $AT$ is most coarse-grained granularity level}
\State $AT_{ot} \gets constructRoot(ot)$
\For{$aot \in sorted_{\text{coarse-to-fine-grained}}(dom(ar))$}
\If{$aot.\text{startsWith}(\text{``c'')})$}
\State $T' \gets \pi_{T}(ON\UP_{ot})$
\If{$adm_T'(ar(aot)$}
\State $AT_{ot} \gets addChild(AT_{ot}, ar(aot), T', ot)$
\EndIf
\Else
\State $TS \gets \{T' \subseteq \pi_{T}(ON\UP_{ot})\,|\,isSESE_{aot}(T') \wedge adm_{T'}(ar(aot))\}$
\For{$T' \in TS$}
\State \Comment{Adding children has to adhere to the control-flow structure}
\State \Comment{hierarchy, e.g., to the tree-structure of process trees.}
\State $AT_{ot} \gets addChildWithHierarchy(AT_{ot}, ar(aot), T', ot, TS)$
\EndFor
\EndIf
\EndFor
\EndFor
\State \Comment{$size$ counts the number of places and transitions of $AN$.}
\While{$size(AN) > sizeThreshold$} 
\State \Comment{$retrieve_{next}$ retrieves abstractions from leafs to root.}
\State $abs, T', ot, aot \gets retrieve_{next}(AT, L) $ 
\State $L \gets st_{abs}(L, T', ot, aot)$ 
\State $AN' \gets overlay(L, AN, ar)$ 
\EndWhile
\EndIf
\State \Return $AN', L, AT$
\end{algorithmic}
\end{algorithm}

$retrieve_{next}$ starts with retrieving complete lifecycle aggregations for lifecycle workflow object types (if present), followed by complete artifact aggregations for subprocess (except the most coarse-grained subprocess), device and resource workflow object types (if present). Then, further aggregations are retrieved from the abstraction tree beginning at the leafs of the IM abstraction trees and switching business object types after each retrieval. Although the occurrence of lifecycle, subprocess, devices and resources workflow object types \janik{leads to} the respective aggregations for lifecycle, subprocess, devices and resources (cf. \autoref{ssec:ot}) \janik{to be added to the abstraction tree at different depth levels of the tree (cf. \autoref{fig:atypes})},  we assume these four workflow object types to be less significant than the business workflow object types as they are either concepts for grouping (subprocess), signify low granularity level (lifecycle), or constitute additional viewpoints on the process (devices, resources). Thus, defining these workflow object types during event extraction has a direct impact on the discovered process model in INEXA.

Each retrieved aggregation $abs$ is applied to the process model through the log-model link. 
Note that the starting point can be an \emph{understandable} process model that is defined as a process model whose size, i.e., number of places and transitions, does not exceed the empirically validated threshold of 37 model elements \cite{sanchez-gonzalez_improving_2013,avila_systematic_2020} or does not significantly exceed the threshold, e.g., $sizeParameter$ in \autoref{alg:init} can be set to 37. Hence, the starting point is determined by abstracting a discovered process model using the abstractions in the initial abstraction repository $ar$ as long as the threshold is exceeded (cf. \autoref{alg:init}). The result is, then, an understandable process model $AN'$ and an augmented event log $L$, in which the abstraction objects and the history abstraction object trace the full sequence of applied aggregations of \texttt{initialize}. Note that the goal of an understandable process model while minimizing information loss can be exchanged with arbitrary other goals such as optimizing performance metrics of the abstracted process model \cite{senderovich_aggregate_2018} to determine the granularity level of the initial process model as a starting point for interactive exploration. As a consequence of a different goal the abstraction tree computation, the condition in the while-loop and the $retrieve_{next}$ functions are potentially affected. 

\setlength{\textfloatsep}{0pt}
\begin{algorithm}[t]
\scriptsize
\caption{Apply}
\label{alg:apply}
\begin{algorithmic}[1]
\Require Augmented event log $L \in \UL$, process model $AN = pd_{IM}(L\UP_{\UWFROT})$, abstraction repository $ar \in \UAOT \rightarrow \UAN \rightarrow \UAN$, abstraction tree $AT$ 
\Ensure Abstracted process model $AN' = abs(overlay(L, AN, ar))$, augmented event log $L' = st_{abs}(L)$, abstraction tree $AT$
\State $AV \gets retrieve_{available}(AT, L)$
\State $abs, T', ot, aot \gets userChoose(AV)$ 
\State $L' \gets st_{abs}(L, T', ot, aot)$ 
\State $AN' \gets overlay(L', AN, ar)$ 
\State \Return $AN', L', AT$
\end{algorithmic}
\end{algorithm}

After the starting point is established through \texttt{initialize}, interactive exploration of a suitable granularity level by the process analyst can proceed with INEXA by means of the \texttt{apply} (cf. \autoref{alg:apply}) and \texttt{redo} (cf. \autoref{alg:redo}) operations. Both are similar in the sense that they first take an augmented event log $L$, e.g., the event log from the starting point, and determine a set of available/redoable aggregations $AV$ from the abstraction tree $AT$ through $retrieve_{available/redoable}$ by means of the currently applied set of aggregations stored in the abstraction history of $L$. This gives the user/process analyst the option to choose from the set of available/redoable aggregations $AV$ ($userChoose$) and to apply/redo the chosen aggregation $abs$ by transitioning the augmented event log with $st_{abs} / st_{abs}^{-1}$ and overlaying the original process model $AN$ with the augmented event log $L'$. 

\setlength{\textfloatsep}{10pt}
\begin{algorithm}[t]
\scriptsize
\caption{Redo}
\label{alg:redo}
\begin{algorithmic}[1]
\Require Augmented event log $L \in \UL$, process model $AN = pd_{IM}(L\UP_{\UWFROT})$, abstraction repository $ar \in \UAOT \rightarrow \UAN \rightarrow \UAN$, abstraction tree $AT$ 
\Ensure Refined process model $AN' = abs^{-1}(overlay(L, AN, ar))$, augmented event log $L' = st_{abs}^{-1}(L)$, abstraction tree $AT$
\State $AV \gets retrieve_{redoable}(AT, L)$
\State $abs, T', ot, aot \gets userChoose(AV)$ 
\State \Comment{$st_{abs}$ is inverted by removing the abstraction object and its history reference. }
\State $L' \gets st^{-1}_{abs}(L, T', ot, aot)$ 
\State $AN' \gets overlay(L', AN, ar)$ 
\State \Return $AN', L', AT$
\end{algorithmic}
\end{algorithm}

\begin{figure}[ht!]
  \centering
  \includegraphics[width=0.85\linewidth]{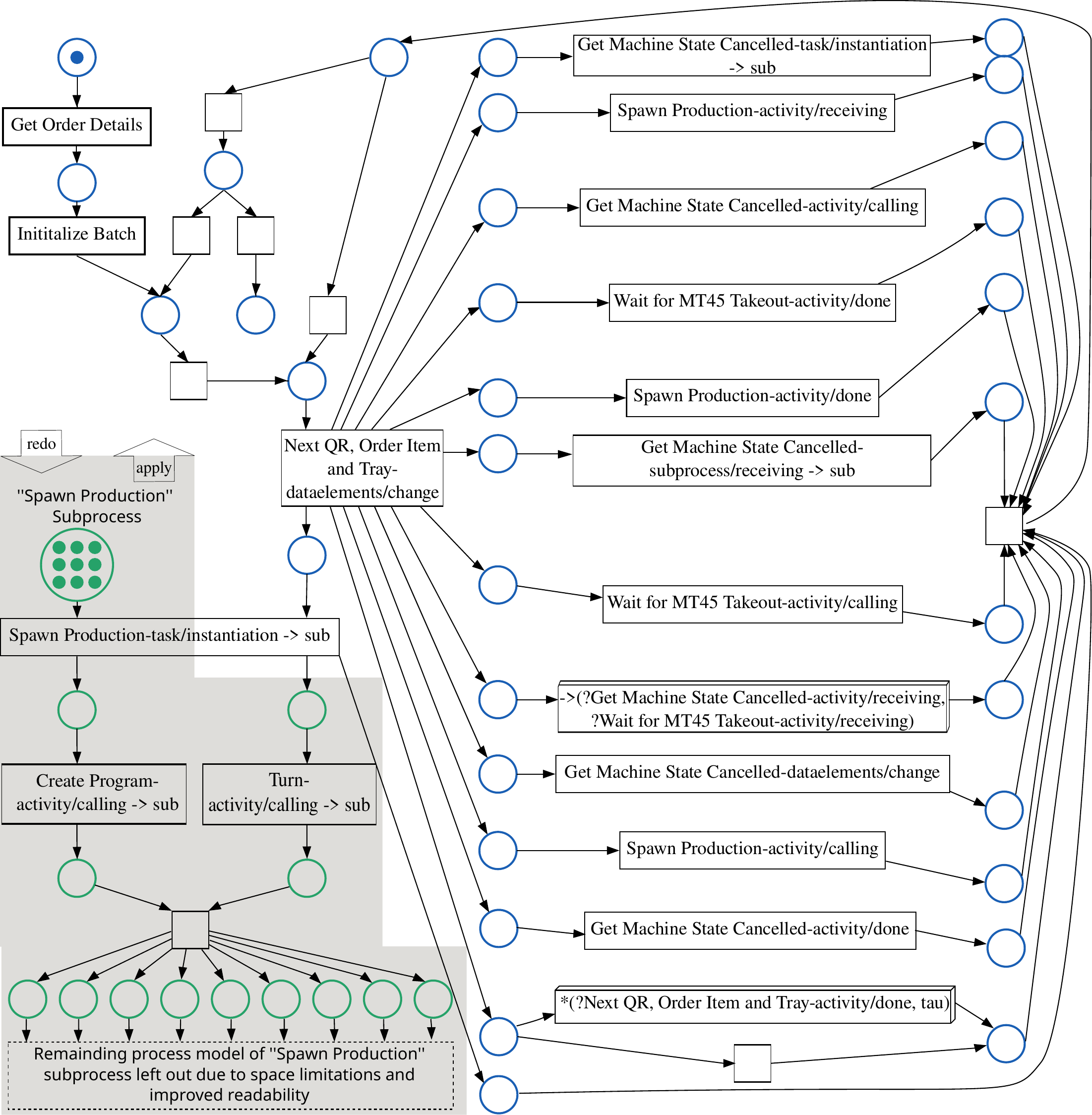}
  \caption{INEXA's initialized process model of the manufacturing process in \cite{mangler_xes_2023}. The initialized process model depicts the orchestration process for producing batches of chess turm pieces. Redoing the ``Spawn Production'' subprocess aggregation shows how the orchestration process spawns the actual production process that calls the respective machine programs to produce the parts.}
  \label{fig:evaluation}
\end{figure}

Differences exist in the functions $retrieve_{available/redoable}$ and in the way a chosen aggregation is handled (applied/redone). While $retrieve_{available}$ determines the next more coarse-grained aggregation in the abstraction tree for each child of the root that corresponds to a business object type from the currently applied aggregations respectively, $retrieve_{redoable}$ determines the most coarse-grained aggregation of the abstraction tree that is currently applied for each child of the root that correspond to a business object respectively. Hence, for each business object, the parent of one of its currently applied aggregations is determined ($retrieve_{available}$) by ``looking upwards'' in the tree from the currently applied nodes or one of the most coarse-grained, currently applied aggregations is determined ($retrieve_{redoable}$). Applying an abstraction in \texttt{apply} works similar to \texttt{initialize}, but redoing a previously applied aggregation means to remove the respective abstraction object and the respective entry in the object attribute $current$ of the abstraction history object through $st_{abs}^{-1}$.

\section{Evaluation}
\label{sec:eval}

We evaluate INEXA that is prototypically implemented in \url{https://github.com/janikbenzin/INEXA} by demonstrating the method on a real-world event log from the manufacturing domain available at \cite{mangler_xes_2023}. The event log consisting of 9.460 events is recorded by the Cloud Process Execution Engine \cite{mangler_cpee_2014} during the execution of the orchestration process that produces nine chess tower pieces in a batch. During production, a MT45 lathe machine\footnote{\url{www.emco-world.com/en/products/turning/maxxturn/maxxturn-45-g2.html}} turns the material to produce the tower. The production of the nine pieces is orchestrated such that production activities for the different pieces are executed in parallel. The chess piece production data is set by the activity ``Next QR, Order Item and Tray-dataelements/change''. Hence, the production cycle starts with this activity followed by a large ``AND'' SESE control-flow structure (cf. \autoref{fig:evaluation}).

\begin{figure}[htb!]
\centering
\caption{Process model metrics for the original model that is discovered without INEXA, the initialized model after INEXA's \texttt{initialize} operation is applied, and the initialized model after the subprocess abstraction for a production subprocess is redone with \texttt{redo}.}
\begin{tabular}{lccc}
 \toprule
 & \makecell{Original Model} & \makecell{Initialized Model} & \makecell{Initialized Model \\ Redo Subpr. Abs.} \\
\midrule
\# \makecell[l]{Model\\Elements}& 1.489 & 58 & 146 \\
\# Arcs & 2.120 & 76 & 182 \\
\# \makecell[l]{Object\\Types} & 89 & 1 & 2 \\
\# \makecell[l]{Sub-\\processes} & 26 & 1 & 2 \\
\bottomrule
\end{tabular}
\label{tab:modelss}
\end{figure}

Given the event log, INEXA' \texttt{initia\-lize} operation with $sizeThreshold$ set to 60\footnote{A size of 37 cannot be achieved with the implemented abstraction repository.} results in the process model in \autoref{fig:evaluation} without the ``Spawn Production'' subprocess denoted in grey. The initialized process model (cf. \autoref{tab:modelss}) is reduced in size by a factor of almost 26 compared to the original process model discovered by OCPD technique $pd_{IM}$ on the event log. The large original process model is the result of the following event log metrics: 245 activity labels with lifecycle suffix, 63 activity labels without lifecycle suffix, 89 object types of which are 26 subprocesses, and 2,118 objects. After the applied ``complete subprocess'' abstraction for the subprocess ``Spawn Production'' is redone with \texttt{redo}, the corresponding subprocess is part of the process model (denoted in grey in \autoref{fig:evaluation}). However, the full process model is not depicted, as the size is almost tripled (cf. \autoref{tab:modelss}). The subprocess ``Spawn production'' shows that a production program is created in parallel to the subprocess start for the MT45 turning. These two activities are followed by another ``AND'' SESE control-flow structure.

Overall, INEXA automatically abstracts the discovered process model until an almost understandable process model of the production orchestration process is the result. The process analyst can explore various granularity levels with the \texttt{apply} and \texttt{redo} operations from this starting point.

\section{Related Work}
\label{sec:rel}

For overviews of event and process model abstraction methods we refer to \cite{van_zelst_event_2021} and \cite{smirnov_business_2012} resp. To the best of our knowledge the main challenge addressed by our research question, interactive and explainable exploration of granularity levels in process models discovered from event logs, has not been addressed yet \cite{beerepoot_biggest_2023}. Since process model abstraction is the interface of our method, we give a short overview of related work in process model abstraction. 
\cite{bobrik_proviado_2006} proposes a visualization model that maintains all visualization parameters required to provide configurable and personalized views on process models. As soon as an appropriate personalized view is configured via XML templates, the view is fixed, i.e., the user interface is rather static than interactive. \cite{senderovich_aggregate_2018} propose a set of process model abstractions to aggregate SESE control-flow structures in \emph{generalized stochastic} Petri nets and state the goal of abstraction (cf. model size goal in \autoref{alg:init}) as an optimization problem of finding the minimal model while guaranteeing a certain error in the performance analysis on the minimal model. As soon as the minimal model is computed given the solution of the optimization problem, no further abstractions are considered. \cite{tsagkani_process_2022} proposes an abstraction method on BPMN models to increase the understandability of the abstracted model compared to the original model. To that end, a set of elimination and aggregation abstractions are defined that not only target SESE control-flow structures, but also lanes and data associations of the BPMN model.

\section{Conclusion and Limitations}
\label{sec:con}

The INEXA method enables interactive exploration of the process model granularity level from a starting point (the result of \texttt{initialize}) by means of the \texttt{apply} and \texttt{redo} operations ($\mapsto$ interactive). Each applied abstraction for reaching the starting point and that is applied/redone during interactive exploration is completely traced in the event log in the form of abstraction objects and the abstraction history through the log-model link. Thus, the full analysis journey recorded in the last augmented event log can be used to explain the resulting process model ($\mapsto$ explainable). Further analysis, e.g., performance analysis, can be conducted either on the non-abstracted process model using the event log $L\UP_{\UWFROT}$ that is kept at the core of INEXA or by defining suitable functions on the domain of the performance analysis that correspond to the respective abstraction, e.g., defining an expected duration for the aggregated transition given an ``Sequence control-flow structure'' abstraction \cite{senderovich_aggregate_2018}, and carrying out the analysis on the aggregated process model. 

INEXA comes with the limitation that the event log has to perfectly fit the originally discovered process model, i.e., the existence of noise and non-conforming business process executions in the event log cannot be handled yet. Considering object-centric alignments \cite{liss_object-centric_2023} is a promising future direction for extending INEXA. Furthermore, the abstraction repository consists of only seven abstractions that all target control-flow structures initially. Also, data recorded with events is not used to define alternative abstraction goals.

\bibliographystyle{splncs04}


\begin{thebibliography}{10}
\providecommand{\url}[1]{\texttt{#1}}
\providecommand{\urlprefix}{URL }
\providecommand{\doi}[1]{https://doi.org/#1}

\bibitem{van_der_aalst_soundness_2011}
van~der Aalst, W.M.P., van Hee, K.M., ter Hofstede, A.H.M., Sidorova~et al.,
  N.: Soundness of workflow nets: classification, decidability, and analysis.
  Form. Asp. Comput.  \textbf{23}(3),  333--363 (2011)

\bibitem{van_der_aalst_process_2016}
van~der Aalst, W.M.P.: Process {Mining}. Springer Berlin Heidelberg (2016)

\bibitem{aalst_replaying_2012}
van~der Aalst, W.M.P., Adriansyah, A., Dongen, B.: Replaying {History} on {Process}
  {Models} for {Conformance} {Checking} and {Performance} {Analysis}. WIREs
  Data Mining and Knowledge Discovery  \textbf{2},  182--192 (Mar 2012)

\bibitem{van_der_aalst_object-centric_2019}
van~der Aalst, W.M.P.: Object-{Centric} {Process} {Mining}: {Dealing} with
  {Divergence} and {Convergence} in {Event} {Data}. In: Ölveczky, P.C.,
  Salaün, G. (eds.) Software {Engineering} and {Formal} {Methods}. pp. 3--25.
  LNCS, Springer International Publishing, Cham (2019)

\bibitem{van_der_aalst_discovering_2020}
van~der Aalst, W.M.P., Berti, A.: Discovering object-centric {Petri} nets.
  Fundamenta informaticae  \textbf{175}(1-4),  1--40 (2020), publisher: IOS
  Press

\bibitem{adams_defining_2022}
Adams, J.N., Schuster, D., Schmitz, S., Schuh, G., van~der Aalst, W.M.P.:
  Defining {Cases} and {Variants} for {Object}-{Centric} {Event} {Data}. In:
  2022 4th {International} {Conference} on {Process} {Mining} ({ICPM}). pp.
  128--135 (Oct 2022)

\bibitem{avila_systematic_2020}
Avila, D.T., dos Santos, R.I., Mendling, J., Thom, L.H.: A systematic
  literature review of process modeling guidelines and their empirical support.
  Business Process Management Journal  \textbf{27}(1),  1--23 (Jan 2020)

\bibitem{beerepoot_biggest_2023}
Beerepoot, I., Di~Ciccio, C., Reijers, H.A., Rinderle-Ma~et al., S.: The
  biggest business process management problems to solve before we die.
  Computers in Industry  \textbf{146},  103837 (Apr 2023)

\bibitem{benzin_preventing_2023}
Benzin, J.V., Park, G., Rinderle-Ma, S.: Preventing {Object}-centric
  {Discovery} of {Unsound} {Process} {Models} for {Object} {Interactions} with
  {Loops} in {Collaborative} {Systems}: {Extended} {Version} (Mar 2023),
  arXiv:2303.16680 [cs]

\bibitem{bergenthum_process_2007}
Bergenthum, R., Desel, J., Lorenz, R., Mauser, S.: Process {Mining} {Based} on
  {Regions} of {Languages}. In: Alonso, G., Dadam, P., Rosemann, M. (eds.)
  Business {Process} {Management}. pp. 375--383. LNCS, Springer, Berlin,
  Heidelberg (2007)

\bibitem{bergenthum_synthesis_2009}
Bergenthum, R., Desel, J., Mauser, S., Lorenz., R.: Synthesis of {Petri} {Nets}
  from {Term} {Based} {Representations} of {Infinite} {Partial} {Languages}.
  Fundamenta Informaticae  \textbf{95}(1),  187--217 (2009)

\bibitem{bertrand_survey_2023}
Bertrand, Y., Van~den Abbeele, B., Veneruso, S., Leotta, F., Mecella, M.,
  Serral, E.: A {Survey} on the {Application} of {Process} {Mining}
  to {Smart} {Spaces} {Data}. In: Montali, M., Senderovich, A., Weidlich, M.
  (eds.) Process {Mining} {Workshops}. pp. 57--70. LNBIP, Springer Nature
  Switzerland, Cham (2023)

\bibitem{bobrik_proviado_2006}
Bobrik, R., Bauer, T., Reichert, M.: Proviado – {Personalized} and
  {Configurable} {Visualizations} of {Business} {Processes}. In: Bauknecht, K.,
  Pröll, B., Werthner, H. (eds.) E-{Commerce} and {Web} {Technologies}. pp.
  61--71. LNCS, Springer (2006)

\bibitem{bobrik_view-based_2007}
Bobrik, R., Reichert, M., Bauer, T.: View-{Based} {Process} {Visualization}.
  In: Alonso, G., Dadam, P., Rosemann, M. (eds.) Business {Process}
  {Management}. pp. 88--95. LNCS, Springer (2007)

\bibitem{buhner_working_2006}
Bühner, M., König, C.J., Pick, M., Krumm, S.: Working {Memory} {Dimensions}
  as {Differential} {Predictors} of the {Speed} and {Error} {Aspect} of
  {Multitasking} {Performance}. Human Performance  \textbf{19}(3),  253--275
  (Jun 2006)

\bibitem{dumas_fundamentals_2013}
Dumas, M., La~Rosa, M., Mendling, J., Reijers, H.A.: Fundamentals of {Business}
  {Process} {Management}. Springer, Berlin, Heidelberg (2013)

\bibitem{ehrendorfer_assessing_2021}
Ehrendorfer, M., Mangler, J., Rinderle-Ma, S.: Assessing the {Impact} of
  {Context} {Data} on {Process} {Outcomes} {During} {Runtime}. In:
  International {Conference} on {Service}-{Oriented} {Computing}. {LNCS}, vol.
  13121, pp. 3--18. Springer (2021)

\bibitem{evans2004domain}
Evans, E., Evans, E.J.: Domain-driven design: tackling complexity in the heart
  of software. Addison-Wesley Professional (2004)

\bibitem{fdhila_verifying_2022}
Fdhila, W., Knuplesch, D., Rinderle-Ma, S., Reichert, M.: Verifying compliance
  in process choreographies: {Foundations}, algorithms, and implementation.
  Information Systems p. 101983 (Jan 2022)

\bibitem{ghahfarokhi_ocel_2021}
Ghahfarokhi, A.F., Park, G., Berti, A., van~der Aalst, W.M.P.: {OCEL}: {A}
  {Standard} for {Object}-{Centric} {Event} {Logs}. In: Bellatreche, L., Dumas,
  M., Karras, P., Matulevičius, R., Awad, A., Weidlich, M., Ivanović, M.,
  Hartig, O. (eds.) New {Trends} in {Database} and {Information} {Systems}. pp.
  169--175. Communications in {Computer} and {Information} {Science}, Springer
  International Publishing, Cham (2021)

\bibitem{kumar_optimal_2013}
Kumar, A., Dijkman, R., Song, M.: Optimal {Resource} {Assignment} in
  {Workflows} for {Maximizing} {Cooperation}. In: Daniel, F., Wang, J., Weber,
  B. (eds.) Business {Process} {Management}. pp. 235--250. Lecture {Notes} in
  {Computer} {Science}, Springer, Berlin, Heidelberg (2013)

\bibitem{leemans_discovering_2013}
Leemans, S.J.J., Fahland, D., van~der Aalst, W.M.P.: Discovering
  {Block}-{Structured} {Process} {Models} from {Event} {Logs} - {A}
  {Constructive} {Approach}. In: Colom, J.M., Desel, J. (eds.) Application and
  {Theory} of {Petri} {Nets} and {Concurrency}. pp. 311--329. LNCS, Springer
  (2013)

\bibitem{reichert_using_2016}
Leemans, S.J.J., Fahland, D., van~der Aalst, W.M.P.: Using {Life} {Cycle}
  {Information} in {Process} {Discovery}. In: Reichert, M., Reijers, H.A.
  (eds.) Business {Process} {Management} {Workshops}, vol.~256, pp. 204--217.
  Springer International Publishing, Cham (2016)

\bibitem{liss_object-centric_2023}
Liss, L., Adams, J.N., van~der Aalst, W.M.P.: Object-{Centric} {Alignments}
  (May 2023), arXiv:2305.05113 [cs]

\bibitem{liu_workflow_2003}
Liu, D.R., Shen, M.: Workflow modeling for virtual processes: an
  order-preserving process-view approach. Information Systems  \textbf{28}(6),
  505--532 (Sep 2003)

\bibitem{mangler_xes_2023}
Mangler, J., Ehrendorfer, M.: {XES} {Chess} {Pieces} {Production} (May 2023).
  \doi{10.5281/zenodo.7958478}

\bibitem{mangler_cpee_2014}
Mangler, J., Rinderle-Ma, S.: {CPEE} - {Cloud} {Process} {Execution} {Engine}.
  In: Limonad, L., Weber, B. (eds.) Proceedings of the {BPM} {Demo} {Sessions}
  2014 {Co}-located with the 12th {International} {Conference} on {Business}
  {Process} {Management} ({BPM} 2014), {Eindhoven}, {The} {Netherlands},
  {September} 10, 2014. {CEUR} {Workshop} {Proceedings}, vol.~1295, p.~51.
  CEUR-WS.org (2014)

\bibitem{park2022detecting}
Park, G., Benzin, J., van~der Aalst, W.M.P.: Detecting context-aware deviations
  in process executions. In: Business Process Management Forum. pp. 190--206
  (2022). \doi{10.1007/978-3-031-16171-1\_12}

\bibitem{pika_mining_2017}
Pika, A., Leyer, M., Wynn, M.T., Fidge, C.J., Hofstede, A.H.M.T., Aalst,
  W.M.P.: Mining {Resource} {Profiles} from {Event} {Logs}. ACM
  Transactions on Management Information Systems  \textbf{8}(1),  1--30 (Mar
  2017)

\bibitem{polyvyanyy2009application}
Polyvyanyy, A., Smirnov, S., Weske, M.: On application of structural
  decomposition for process model abstraction. Business Process, Services
  Computing and Intelligent Service Management: BPSC 2009 pp. 110--122 (2009)

\bibitem{polyvyanyy_simplified_2011}
Polyvyanyy, A., Vanhatalo, J., Völzer, H.: Simplified {Computation} and
  {Generalization} of the {Refined} {Process} {Structure} {Tree}. In: Bravetti,
  M., Bultan, T. (eds.) Web {Services} and {Formal} {Methods}. pp. 25--41.
  LNCS, Springer, Berlin, Heidelberg (2011)

\bibitem{reijers_ha_usefulness_2010}
{Reijers H.A.}, {Mendling J.}, {Dijkman R.M.}: On the usefulness of
  subprocesses in business process models. BPM reports  \textbf{1003} (2010)

\bibitem{russell2004workflow}
Russell, N., ter Hofstede, A., Edmond, D., van~der Aalst, W.M.P.: Workflow resource
  patterns. BETA publicatie: working papers  \textbf{127} (2004)

\bibitem{senderovich_aggregate_2018}
Senderovich, A., Shleyfman, A., Weidlich, M., Gal, A., Mandelbaum, A.: To
  aggregate or to eliminate? {Optimal} model simplification for improved
  process performance prediction. Information Systems  \textbf{78},  96--111
  (Nov 2018)

\bibitem{smirnov_business_2012}
Smirnov, S., Reijers, H.A., Weske, M., Nugteren, T.: Business process model
  abstraction: a definition, catalog, and survey. Distributed and Parallel
  Databases  \textbf{30}(1),  63--99 (Feb 2012)

\bibitem{smirnov2012business}
Smirnov, S., Weidlich, M., Mendling, J.: Business process model abstraction
  based on synthesis from well-structured behavioral profiles. International
  journal of cooperative information systems  \textbf{21}(01),  55--83 (2012)

\bibitem{sanchez-gonzalez_improving_2013}
Sánchez-González, L., Ruiz, F., García, F., Piattini, M.: Improving
  {Quality} of {Business} {Process} {Models}. In: Maciaszek, L.A., Zhang, K.
  (eds.) Evaluation of {Novel} {Approaches} to {Software} {Engineering}. pp.
  130--144. Communications in {Computer} and {Information} {Science}, Springer,
  Berlin, Heidelberg (2013)

\bibitem{tsagkani_process_2022}
Tsagkani, C., Tsalgatidou, A.: Process model abstraction for rapid
  comprehension of complex business processes. Information Systems
  \textbf{103},  101818 (Jan 2022)

\bibitem{turetken_influence_2020}
Turetken, O., Dikici, A., Vanderfeesten, I., Rompen~et al., T.: The {Influence}
  of {Using} {Collapsed} {Sub}-processes and {Groups} on the
  {Understandability} of {Business} {Process} {Models}. Business \& Information
  Systems Engineering  \textbf{62}(2),  121--141 (Apr 2020)

\bibitem{vanhatalo_refined_2009}
Vanhatalo, J., Völzer, H., Koehler, J.: The refined process structure tree.
  Data \& Knowledge Engineering  \textbf{68}(9),  793--818 (Sep 2009)

\bibitem{van_zelst_translating_2020}
van Zelst, S.J., Leemans, S.J.J.: Translating {Workflow} {Nets} to {Process}
  {Trees}: {An} {Algorithmic} {Approach}. Algorithms  \textbf{13}(11), ~279
  (Nov 2020)

\bibitem{van_zelst_event_2021}
van Zelst, S.J., Mannhardt, F., de~Leoni, M., Koschmider, A.: Event abstraction
  in process mining: literature review and taxonomy. Granular Computing
  \textbf{6}(3),  719--736 (Jul 2021)

\bibitem{zerbato_granularity_2021}
Zerbato, F., Seiger, R., Di~Federico, G., Burattin, A., Weber, B.: Granularity
  in {Process} {Mining}: {Can} we fix it? In: {CEUR} {Workshop} {Proceedings}.
  vol.~2938, pp. 40--44 (2021)

\end{thebibliography}
\end{document}